# Refinement-Cut: User-Guided Segmentation Algorithm for Translational Science

Jan Egger

*Faculty of Computer Science and Biomedical Engineering*
*Institute for Computer Graphics and Vision*
*Graz University of Technology, Graz, Styria, Austria.*

egger@tugraz.at

**Abstract**

In this contribution, a semi-automatic segmentation algorithm for (medical) image analysis is presented. More precise, the approach belongs to the category of interactive contouring algorithms, which provide real-time feedback of the segmentation result. However, even with interactive real-time contouring approaches there are always cases where the user cannot find a satisfying segmentation, e.g. due to homogeneous appearances between the object and the background, or noise inside the object. For these difficult cases the algorithm still needs additional user support. However, this additional user support should be intuitive and rapid integrated into the segmentation process, without breaking the interactive real-time segmentation feedback. I propose a solution where the user can support the algorithm by an easy and fast placement of one or more seed points to guide the algorithm to a satisfying segmentation result also in difficult cases. These additional seed(s) restrict(s) the calculation of the segmentation for the algorithm, but at the same time, still enable to continue with the interactive real-time feedback segmentation. For a practical and genuine application in translational science, the approach has been tested on medical data from the clinical routine in 2D and 3D.

**Introduction**

Nowadays, the most clinics – at least in the western world – have in general several medical scanners, like computed tomography (CT) or magnetic resonance imaging (MRI), which produce every day a massive amount of medical patient data. In addition, new scanner generations get more and more precise, and thus produce more and more data. However, there is by far not the time and manpower for a precise manual analysis of this important and critical data. Therefore, approximations are often used, like the estimated calculation of a tumor volume via its maximal diameter in a 2D view, which may not very accurate and can lead to inaccurate treatment decisions[1]. A solution could be to support and automate medical image analysis with segmentation algorithms, like Active Contours in 2D[2] or 3D[3], Active Appearance Models[4], graph-based approaches[5], fuzzy-based approaches[6], or neural networks[7]. But after observing dozen of interventions in several clinics and different departments, I never met a physician who used any segmentation algorithm. The main reason was, that the segmentation approaches are not stable enough and fail far too often, especially for fully automatic algorithms. This may also be the reason that major manufacturers of medical imaging equipment don't really offer sophisticated segmentation options within their workstations and software packages. Additionally, the existing approaches are often not user friendly and intuitive implemented, e.g. they need a precise definition of "mystic" parameters for an accurate segmentation result. A temporal solution to speed-up a segmentation task, until (fully) automatic algorithms provide reliable results, are semi-automatic methods, like interactive segmentation approaches. Thereby, the user supports and guides the algorithm by interactive input. This can be carried out by marking parts of the pathology and the surrounding background with a simple brush[8]. Thus providing the segmentation algorithm with information about the pathology's location in the image and information about the texture of the pathology and background[9,10]. An overview about several interactive medical image segmentation approaches has recently been published by Zhao and Xie X[11], where they also classify the approaches by their type of interactions:

- Pictorial input on an image grid, like Seeds for region growing[12],
- Parameter tuning using slider, dial, or similar interface, like the maximum size of segmented regions[13], or
- Menu option selection by mouse clicking, like Accept/reject the segmentation results[14].

An exciting (new) class of interactive segmentation algorithms – which are not discussed to detailed within the review – are real-time approaches, which are able to calculate a segmentation result within a fraction of a second. In the meantime, this is possible because hardware becomes faster and faster and therefore allows the execution of high level segmentation approaches in an extremely short time, even on up-to-date laptops. This opens up completely new possibilities, where the user gets immediate feedback, instead of waiting for the segmentation result to come back and then re-initialize and start over again, which can be very frustrating. A real-time interactive image segmentation approach that uses user indicated real-world seeds has been presented by Gomes et al.[15]. The approach can be used for videos or still images and because the seeds are indicated by a user, e.g. via a laser pointer, it is possible to segment objects without any computer interface. Armstrong et al.[16] introduce interactive segmentation of image volumes with live surface. In summary, Live Surface does for 3D volumes what Intelligent Scissors[17,18] did for 2D images, and allows the user to segment volumes continuously with immediate visual feedback in the refinement of the selected surface. A variational model for interactive shape prior segmentation and real-time tracking has been proposed by Werlberger et al.[19]. The semi-automated segmentation approach is based on minimizing the Geodesic Active Contour[20] energy incorporating a shape prior that represents the desired structure. Additionally, the user has the possibility to make corrective during the segmentation and adapt the shape prior position. To achieve a real-time behavior the method was implemented on the GPU. A computer-aided design system for refinement of segmentation errors has been introduced by Jackowski and Goshtasby[21], where a surface is interactively revised until the desired segmentation has been achieved. Therefore, the surface is revised by moving certain control points and the user sees the changes in the surface in real-time. Mory et al.[22] propose a real-time 3D image segmentation method based on user-constrained template deformation. The interactive image segmentation algorithm incorporates in a first step user input as inside/outside labeled points to drive the deformation and improve both robustness and accuracy. In a second step, a fast implementation of non-rigid template-to-image registration enables interactions with a real-time visual feedback.

In this contribution, an interactive real-time segmentation algorithm is introduced. The algorithm is scale-invariant and keeps its interactive real-time segmentation behavior even if the user refines the segmentation result with additional seeds. Thus, in principle, the algorithm combines some basic characteristics from existing segmentation methods into a novel segmentation approach which can also handle difficult segmentation task; and to the best of the author's knowledge such an approach has not yet been described.

The paper is organized as follows: The Materials and Methods section presents the details of the proposed algorithm and online resources where medical data can be found; the Results section displays the outcomes of my experiments; and the Discussion section concludes the paper and outlines areas for future research.

**Results**

Figure 1 presents the interactive refinement segmentation of a vertebral body contour in 2D from a MRI acquisition. The leftmost image shows the native scan and the second image from the left displays the initial user-defined seed point (white) that has been placed inside the vertebral body for the interactive segmentation. The third image from the left presents the segmentation outcome for the current position of the user-defined seed. However, due to the bright region inside the vertebral body, the average gray value – which is automatically calculated from a region around the user-defined seed – is not detected "correctly", and thus the resulting contour (red) leaks in the upper area and misses an edge in the lower left (note: for the interactive segmentation of the vertebral body, a rectangle was used as template to construct the graph. Thereby, the center of the rectangle is at the position of the user-defined seed point and the yellow crosses in the two rightmost images display the four corners of the rectangle). Nevertheless, the rightmost image presents the result of a refined segmentation. Therefore, the user simply placed three additional seeds (white dots on the contour of the vertebral body), and thus forced the algorithm to perform the min-cut at these positions – which also influences the cuts along the neighboring rays. Furthermore, additional gray value information can be extracted around these extra seeds that the user placed on the contour of the vertebral body.

Figure 2 presents the interactive refinement segmentation of the rectum from an intraoperative gynecological 3-Tesla magnetic resonance imaging dataset. The leftmost image shows the native scan and the second image from the left presents the initial seed point (white) for the interactive segmentation placed by the user inside the rectum. The red dots present the segmentation outcome with regard to the current seed point position (note: for the interactive segmentation of the rectum, a triangle was used as template to construct the graph. Thereby, the center of the rectangle is located at the user-defined seed point and the yellow crosses display the three corners of the triangle). In the third image from the left an additional seed point (white) has been placed in the upper left contour of the rectum. This additional seed forces the algorithm to perform the min-cut at this position. In the fourth image from the left, the user has interactively repositioned the initial seed point inside the rectum to find a better segmentation outcome. However, the additional seed at the contour stays fixed during the interactive repositioning of the initial seed and still forces the algorithm to perform the min-cut at its position in the upper left contour of the rectum. In the rightmost image, the user further refined the segmentation outcome with two additional seed points.

Figure 3 presents the interactive segmentation of a stented lumen and the thrombus from a postoperative computed tomography angiography (CTA) scan from a patient with an abdominal aortic aneurysm (AAA)[23,24]. The leftmost image shows the original scan and the second image from the left presents the segmentation of the stented lumen (red) with the initial user-defined seed point (green) placed inside the lumen (note: for the interactive segmentation a circle was used as template to construct the graph). The following three images show how the user places a second seed point and interactively drags it to the contour of the thrombus. However, the graph is still constructed from the initial seed point that has been placed at first inside the lumen. The second seed point forces the algorithm to perform the min-cut at its position and therefore also influences the positions of the min-cut in the neighboring rays. During the interactive dragging of the second seed inside the thrombus (images three and four from the left), the algorithm tries to adapt to other structures appearing in the thrombus. In this example, contrast enhanced blood from an endoleak[25] is visible (elongated bright area inside the thrombus), and the resulting contour partly fits to this endoleak in the third and the fourth image in the lower right area: once to the left contour of the endoleak (third image) and once to the right contour of the endoleak (fourth image). In the rightmost image, the segmentation outcome has furthermore been refined by an additional seed point placed by the user on the contour of the thrombus in the lower left.

Figure 4 presents the interactive segmentation of the prostate central gland (PCG) in 3D with a spherical template. The leftmost images show the original scan in axial (top), coronal (middle) and sagittal (bottom) views. The second image from the left presents the segmentation outcome (red) for a user-defined seed point (blue) placed inside the prostate (note: the seed point has been placed in the axial view, even if it is also displayed in the coronal and sagittal views). For comparison, the green masks display the outcome of a manual slice-by-slice segmentation from an expert. However, as the initial seed point is placed close to the right border of the prostate, the algorithm missed the contours on the left side of the PCG (axial and coronal views). Though, the interactive real-time behavior of the approach makes a repositioning easy, and thus it is also easy to find a good segmentation outcome for the axial, coronal and sagittal views (third image from the left). In the rightmost image, the segmentation result has been further refined with an additional seed that has been placed by the user in the lower right within the sagittal view.

Figure 5 presents different views – axial (top), coronal (middle) and sagittal (bottom) – of the 3D segmentation outcome from Figure 4. The left images show the last nodes (red) that still belong to the foreground (PCG) after the min-cut, and therefore defining the prostate central gland. In the images displayed in the middle column, the segmentation result has been superimposed with the manual mask (green) from the slice-by-slice expert segmentation. Finally, the rightmost images present a closed surface form the graph's nodes, which can be used to generate a solid mask of the segmentation outcome for further processing.

In addition, performance tests have been carried out with a square template for vertebral body segmentation[26] on a laptop with Intel Core i5-750 CPU, 4 × 2.66 GHz, 8 GB RAM running Windows 7 Professional x64 Version. Thereby, the computation time included the graph construction (sending out the rays from the user-defined seed point, sampling the nodes along these rays and constructing the edges), analyzing the average gray value around the user-defined seed point (which is incorporated into weights of the graph's edges) and the optimal mincut calculation to separate the background from the foreground. The diameter of the square template was set to 80 mm and the delta value was set to 2. For 900 nodes (coming from 30 rays and 30 nodes per ray), an average interactive segmentation time of 30 ms could be achieved. For 9.000 nodes (300 rays, 30 nodes per ray), the segmentation time was in general still under 100 ms, which is still acceptable and within the time range from current smartphone touchscreens[27].

However, for 90.000 nodes (3.000 rays and 30 nodes per ray, or 300 rays and 300 nodes per ray) the average time was around 130 ms, where a minor latency time could already been recognized. That would mean the approach is not real-time anymore, but from a user point of view this is still acceptable for an interactive segmentation process. In contrast, 900.000 nodes (30.000 rays, 30 nodes per ray) were too slow for a convenient interactive segmentation, because the computation time went up to one second.

The outcome of the final segmentations for the presented interactive approach is heavily dependent on the manually placed seed points. However, in previous publications the segmentation of medical pathologies (like Glioblastoma Multiforme, Pituitary Adenomas, Cerebral Aneurysms, Prostate Central Glands and Vertebral Bodies) have already been evaluated via one fixed user-defined seed point, and the summary of these results have been presented here[28]. There, it could already show that a DSC around 80% is possible with only one seed point. However, in principle a user can get very close to the ground truth (manual segmentation) if enough manual seed points are added. Figure 6 presents an example of the prostate where several seed points (white) have been placed to get a segmentation result (red) that matches almost perfect with the manual segmentation (green).

**Discussion**

In this study, an interactive contouring algorithm for image segmentation, with a strong focus on medical data, has been introduced. More specific, the presented algorithm belongs to the class of interactive contouring approaches, which provide immediate feedback of the segmentation result to the user. Thus, allowing the user to interfere easily and intuitive into the algorithms calculation of the segmentation result. Nevertheless, there are always cases where the user cannot find a satisfying segmentation, when an algorithm has to detect the majority of the objects contour. This can have several reasons, the most frequent are in general homogeneous appearances between the object and the background, noise within the object to segment, or "complex" shapes of the object. For these difficult cases an algorithm requires additional support. This support should be intuitive and fast accomplishable by the user, and furthermore allows to continue the interactive segmentation. The proposed solution in this contribution is an easy and fast interactive placement of additional seed points in case of an unsatisfying segmentation outcome. Moreover, the approach allows to come back to an interactive refinement of the initial seed point, even under the new restrictions of the additional seeds. Furthermore, the additional seeds can provide the algorithm with broader geometrical and textural information and therefore restrict the possible segmentation calculation even more. For an initial feasibility evaluation, the approach has been implemented within a medical prototyping platform and tested mainly two- and three-dimensional medical data from the clinical routine, with the ultimate goal to assist pure manual slice-by-slice outlining.

The novelty within this study lies in the combination of several pre-developed segmentation techniques[26,28-32], resulting in an advanced interactive *real-time* contouring algorithm for (medical) data. More specific, the presented work extends and incorporates a refinement option[29,30] – introduced only for fixed seed points and a spherical shape[31] – into the recently published Interactive-Cut[28] algorithm that can handle arbitrary shapes[32], but had no refinement option. In sum, the achieved research highlights of the study are:

- An novel interactive contouring algorithm has been designed;
- The algorithm combines shape-based segmentation with user refinement;
- The user refinement is intuitive and fast, with immediate feedback;
- The segmentation works on 2D and 3D image data;
- The evaluation has been performed on medical data from the clinical routine.

There are several areas for future work: in particular, supporting manual strokes from the user which have been drawn along the border of the object to segment – instead of "only" single seed points. Albeit, this may "break" the real-time feedback you get from single seed points even if these are dragged on the image. As shown in the result section, a single seed point can still be moved around to find a better segmentation result, this is not so easy and intuitive anymore if the user has once drawn a stroke. Though, a solution may be an iterative adaption to the manually sketched parts of the user[33].

Furthermore, a detailed study for the end user (which are primarily physicians in case of medical data) is necessary. Even if several physicians from different fields already tested the approach and responded positively, it's of course not certain that they will use it for own research (e.g. for a time-consuming analysis of medical data for own research purpose) or even in the clinical routine. However, after carrying out several studies with a typical stroke-based approach[9,10,34], it is clear that they would only accept such a course of action (the initialization) if the segmentation outcome is afterwards always satisfying (note: for the automatic segmentation the participating physicians had only to mark parts of the fore- and background with a simple brush; no other settings or parameters had to be defined). However, a long-term end user study regarding the presented approach already has been started within two European funded projects ClinicIMPPACT ([www.clinicimppact.eu/](www.clinicimppact.eu/)) and GoSmart ([www.gosmart-project.eu](www.gosmart-project.eu)), where post-interventional radiofrequency ablation (RFA) zones are segmented[35].

**Materials and Methods**

*Data* – For a practical and genuine application in translational science, the elaborated approach has been tested with two-dimensional and three-dimensional medical data from the clinical routine. Intraoperative gynecological 3-Tesla magnetic resonance imaging datasets that have been used for this study can be found here[36,37,38]. MRI datasets of the spine, which are public available for research purposes, can be downloaded here[39,40,41].

Pre- and intra-procedural MR-guided prostate biopsy datasets with manual segmentations are freely available here[42,43,44].

*Software* – The presented approach has been implemented as own C++ module within the medical prototyping platform MeVisLab (www.mevislab.de, Version 2.3, Date of access: 28/04/2014) under the 64-bit version of Windows 7 Professional. Thereby, basic functionalities provided by MeVisLab, like loading medical data, e.g. in the DICOM format (OpenImage module), viewing and navigating through 2D slices (View2D module), displaying data and results in 3D (View3D module) and placing seed points (SoView2DMarkerEditor module) have been used. To calculate the max-flow/min-cut on graphs, the public available source code from Yuri Boykov and Vladimir Kolmogorov has been used (http://vision.csd.uwo.ca/code/, Version 3, Date of access: 28/04/2014)[45]:

*Algorithm* – The core algorithm has been implemented as own MeVisLab C++ module and is a combination and extension of the Template-Cut[32] and the Interactive-Cut[28] approaches, and the refinement method introduced in[29,30]. The new algorithm (Refinement-Cut), as well as the predecessor methods it builds up, belong to the graph-based approaches. Here, an image is interpreted as graph $G(V,E)$ which consists of nodes $n \in V$ sampled in the image and edges $e \in E$ establishing connections between nodes. After graph construction a minimal s-t-cut[45] is calculated on the graph, dividing the nodes into two disjoint sets, whereby one set the segmented objects and one set the background represents – note: for the calculation of the minimal s-t-cut, two additional virtual nodes $s \in V$ (called *source*) and $t \in V$ (called *sink*) are used. The minimal s-t-cut returns the global optimum on a constructed graph, in contrast to iterative approaches, like the Active Contours, which in general find a solution stepwise, and thus can get stuck during this process in a local minimum. However, the immediate calculation of a global optimum, like the minimal s-t-cut, makes graph-based approaches in particular eligible for an interactive real-time application. First of all, for the graph construction, the nodes $n \in V$ are sampled along rays which are sent out from one single seed point and with regards to a certain template. This template represents the basic shape of the segmented object, like described in the Template-Cut approach. Examples are

- A rectangle shape for vertebra segmentation in 2D[26];
- A circle template for prostate central gland segmentation in 2D[28];
- A cubic shape for vertebral body segmentation in 3D[46,47];
- A spherical shape for prostate central gland or brain tumor segmentation in 3D[31,43];
- Or even a user-defined shape for objects that vary too much to be predefined by a simple shape[48].

After the nodes and the underlying texture values within the image have been sampled, the graph's edges $E$ are generated, that establish the connections between the (virtual) nodes, and an edge $\langle v_i, v_j \rangle \in E$ defines the connection between the two nodes $v_i, v_j$. Taking over the notation of Li et al.[49], there are two types of $\infty$-weighted edges:

- Intra-edges which connect nodes along the same ray to ensure that the minimal s-t-cut runs through only one edge within this ray;

- Inter-edges which connect nodes from different rays under a smoothness value delta $\Delta_r$, which influences the number of possible s-t-cuts and therefore the flexibility of the resulting segmentation.

Furthermore, there are edges between the sampled nodes and the virtual nodes ($s$ and $t$) for the graph construction established, and the weights of these edges depend on the sampled texture values within the image and a cost function. For more detail about the graph construction the reader is referred at this point to the previous Template-Cut publication[32]. However, the specific graph construction, which basically starts from one single seed point inside the segmentation object, is particularly suitable for an interactive real-time segmentation, because the user has only to drag this one single seed point over the image – in contrast, to approaches where more input like information about fore- and background or strokes are needed. Moreover, the user can easily add more seed points on the object's contour, which modify the graph and force the minimal s-t-cut to go through this additional seeds. Thereto, the algorithm search for the graph's node that is closest to the additional seed point provided by the user (note: In general, the additional seed's position will not match 100% with the position of a sample node, especially for a low density of rays and sampled nodes, rather the closest graph's node $c$ is chosen). In a next step, the minimal s-t-cut has to be forced to be at this position. In order to ensure that, the graph's node $c$ and all its predecessors within the same ray are connected via $\infty$-weighted edges to the source $s$, and all successor of $c$ within the same ray are connected via $\infty$-weighted edges to the sink $t$. Furthermore, the intra-edge between $c$ and its direct successor node within the same ray is removed. That this course of action works, has already been shown in an initial study with fixed seeds point for the segmentation of glioblastoma multiforme (GBM)[30], where the Dice Similarity Score (DSC)[50] could be improved from 77.72% to 83.91%. However, the possibility to drag an additional seed around an image and at the same time getting the updated segmentation result, makes this approach much more powerful and therefore the finding of a satisfying segmentation result much more convenient. Nevertheless, during dragging the closest graph's node $c$ will most likely change, and has to be re-calculated as soon as the graph is re-constructed. But this allows the user to drag the additional seed points to arbitrary positions on the image and even works if a seed point is outside the predefined template. The additional user-defined seed points also influence the position of the minimal s-t-cuts on the neighboring rays. This influence gets even stronger for lower delta values, which restricts the flexibility of the resulting segmentation. Hence, there are many things going on "under the hood" (and hidden for the user) but still have to be handled in real-time during the interactive dragging of the seeds on the image.

**Acknowledgements**

Primarily, the author thanks the physicians Dr. med. Barbara Carl, Thomas Dukatz, Christoph Kappus, Dr. med. Malgorzata Kolodziej, Dr. med. Daniela Kuhnt and Professor Dr. med. Christopher Nimsky for providing the neurosurgical datasets, and performing manual slice-by-slice segmentations for the evaluation of the algorithm. The author also thanks Drs. Fedorov, Tuncali, Fennessy and Tempany, for sharing the prostate data collection plus manual segmentations, and Fraunhofer MeVis for providing an academic MeVisLab license. The work received funding from the European Union in FP7: Clinical Intervention Modelling, Planning and Proof for Ablation Cancer Treatment (ClinicIMPPACT, grant agreement no. 610886) and Generic Open-end Simulation

Environment for Minimally Invasive Cancer Treatment (GoSmart, grant agreement no. 600641).

**Additional Information**

**Author contribution statement**

Conceived and designed the experiments: JE. Performed the experiments: JE. Analyzed the data: JE. Contributed reagents/materials/analysis tools: JE. Wrote the paper: JE.

**Competing financial interest statement**

The author in this paper has no competing financial interests.

**Figure Legends**

**Figure 1** – Interactive refinement segmentation of a vertebral body contour in 2D from a magnetic resonance imaging (MRI) acquisition. The leftmost image presents the native scan and the second image from the left shows the initial user-defined seed point (white) that has been placed inside the vertebral body for the interactive segmentation. The third image from the left presents the segmentation outcome for the current position of the user-defined seed. However, due to the bright region inside the vertebral body, the average gray value – which is automatically calculated from the region around the user-defined seed – is not calculated "correctly", and thus the resulting contour (red) leaks in the upper area and misses an edge in the lower left (note: for the interactive segmentation

of the vertebral body, a rectangle was used as template to construct the graph. Thereby, the center of the rectangle is the user-defined seed point and the yellow crosses in the two rightmost images display the four corners of the rectangle). Finally, the rightmost image presents the result of the refined segmentation. Therefore, the user simply placed three additional seeds (white dots on the contour of the vertebral body), and thus forced the algorithm to perform the min-cut at these positions – which also influences the cuts along the neighboring rays. Furthermore, additional gray value information can be extracted around these extra seeds that the user placed on the contour of the vertebral body.

**Figure 2** – Interactive refinement segmentation of the rectum from an intraoperative gynecological 3-Tesla magnetic resonance imaging dataset. The leftmost image shows the native scan and the second image from the left presents the initial seed point (white) for the interactive segmentation placed by the user inside the rectum. The red dots present the segmentation outcome with regard to the current seed point position (note: for the interactive segmentation of the rectum, a triangle was used as template to construct the graph. Thereby, the center of the rectangle is the user-defined seed point and the yellow crosses display the three corners of the triangle). In the third image from the left, an additional seed point (white) has been placed in the upper left contour of the rectum. This additional seed forces the algorithm to perform the min-cut at this position. In the fourth image from the left, the user has interactively repositioned the initial seed point inside the rectum to find a better segmentation outcome. However, the additional seed at the contour stays fixed during the interactive repositioning of the initial seed and still forces the algorithm to perform the min-cut at its position in the upper left contour of the rectum. In the rightmost image, the user further refined the segmentation outcome with two additional seed points.

**Figure 3** – Interactive segmentation of a stented lumen and the thrombus from a postoperative computed tomography angiography (CTA) scan from a patient with an abdominal aortic aneurysm (AAA). The leftmost image shows the original scan and the second image from the left presents the segmentation of the stented lumen (red) with the initial user-defined seed point (green) that has been placed inside the lumen (note: for the interactive segmentation a circle was used as template to construct the graph). The following three images show how the user places a second seed point and interactively drags it to the contour of the thrombus. However, the graph is still constructed from the initial seed point that has been placed inside the lumen. In addition, the second seed point forces the algorithm to perform the min-cut at its position and therefore also influences the positions of the min-cut in the neighboring rays. During the interactive dragging of the second seed point inside the thrombus (image three and image four from the left), the algorithm tries to adapt to other structures visible in the thrombus. In this example, contrast enhanced blood from an endoleak is visible (elongated bright area inside the thrombus), and the resulting contour adapts to this endoleak in the third and the fourth image in the lower right area: once to the left contour of the endoleak (third image) and once to the right contour of the endoleak (fourth image). In the rightmost image, the segmentation outcome has been furthermore refined by an additional seed point placed on the contour of the thrombus in the lower left.

**Figure 4** – Interactive segmentation of the prostate central gland (PCG) in 3D with a spherical template. The leftmost images show the original scan in axial (top), coronal (middle) and sagittal (bottom) views. The second image from the left presents the segmentation outcome (red) for a user-defined seed point (blue) that has been placed inside the prostate (note: the seed point has been placed in the axial view, but it is also displayed in the coronal and sagittal views). For comparison, the green masks display the outcome of a manual slice-by-slice segmentation from an expert. However, as the initial seed point is placed close to the right border of the prostate, the algorithm missed the contours of the PCG on the left (axial and coronal views). Though, the interactive real-time behavior of the approach makes a repositioning easy, and thus it is also easy finding a good segmentation outcome for the axial, coronal and sagittal views (third image from the left). In the rightmost image, the segmentation result has been further refined with an additional seed that has been placed in the lower right within the sagittal view.

**Figure 5** – Different views – axial (top), coronal (middle) and sagittal (bottom) – of the 3D segmentation outcome from Figure 4. The left images present the last nodes (red) that still belong to the foreground after the min-cut, and therefore they define the segmented prostate central gland contour. In the images of the middle column, the segmentation result has been superimposed with the manual mask (green) from the slice-by-slice expert segmentation. Finally, the rightmost images present a closed surface form the graph's nodes, which can be used to generate a solid mask of the segmentation outcome for further processing.

**Figure 6** – Semi-automatic segmentation of the prostate where several seed points (white) have been placed to get a segmentation result (red) that matches almost perfect with a pure manual segmentation (green).

# **Figures**

Figure 1

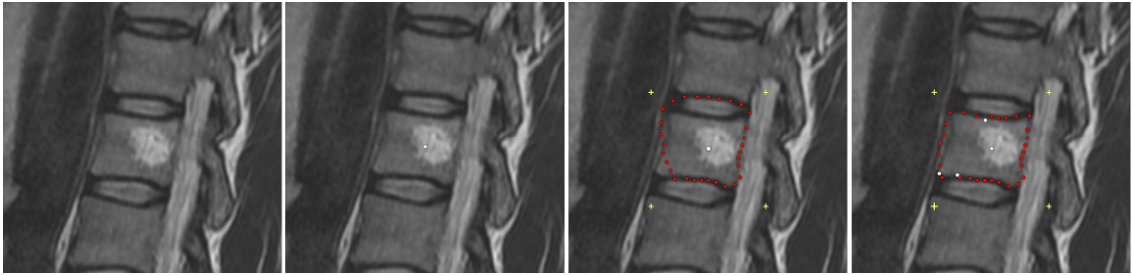

Figure 2

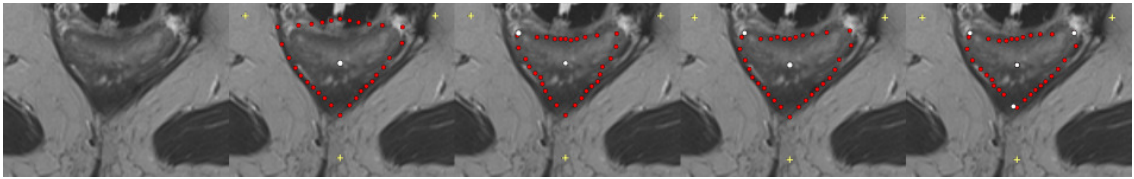

Figure 3

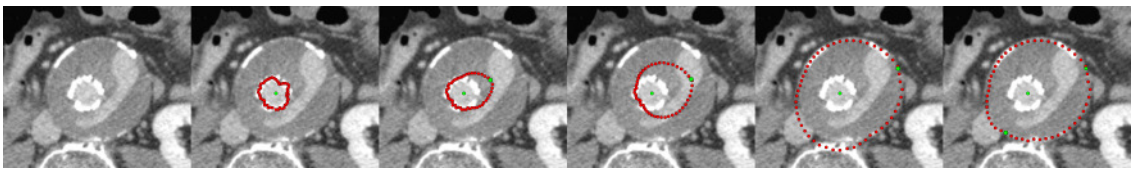

Figure 4

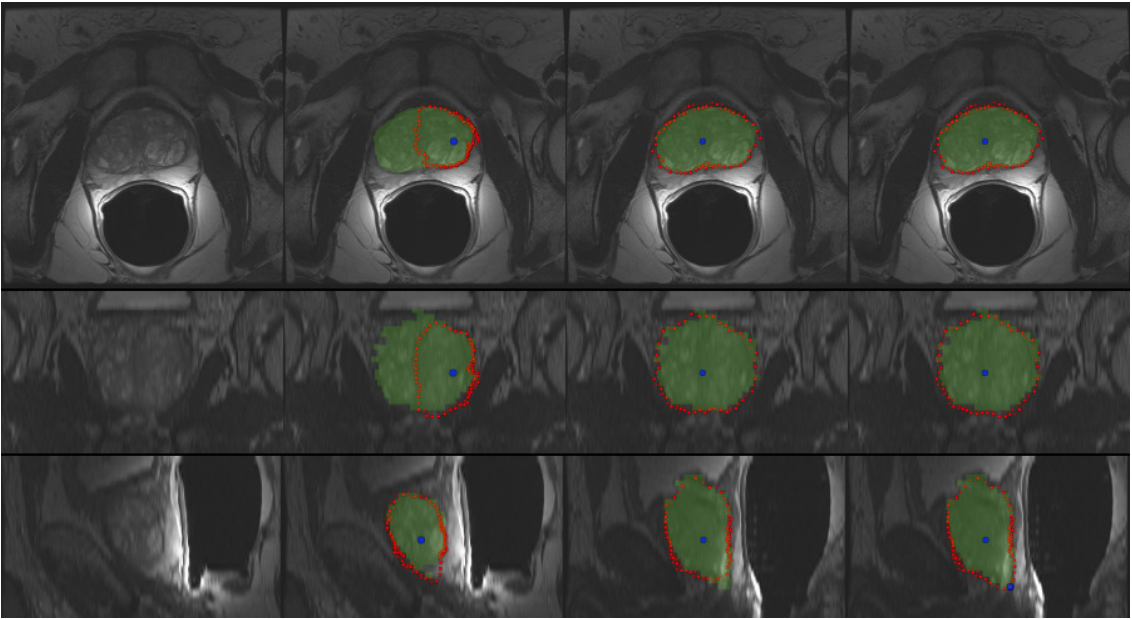

Figure 5

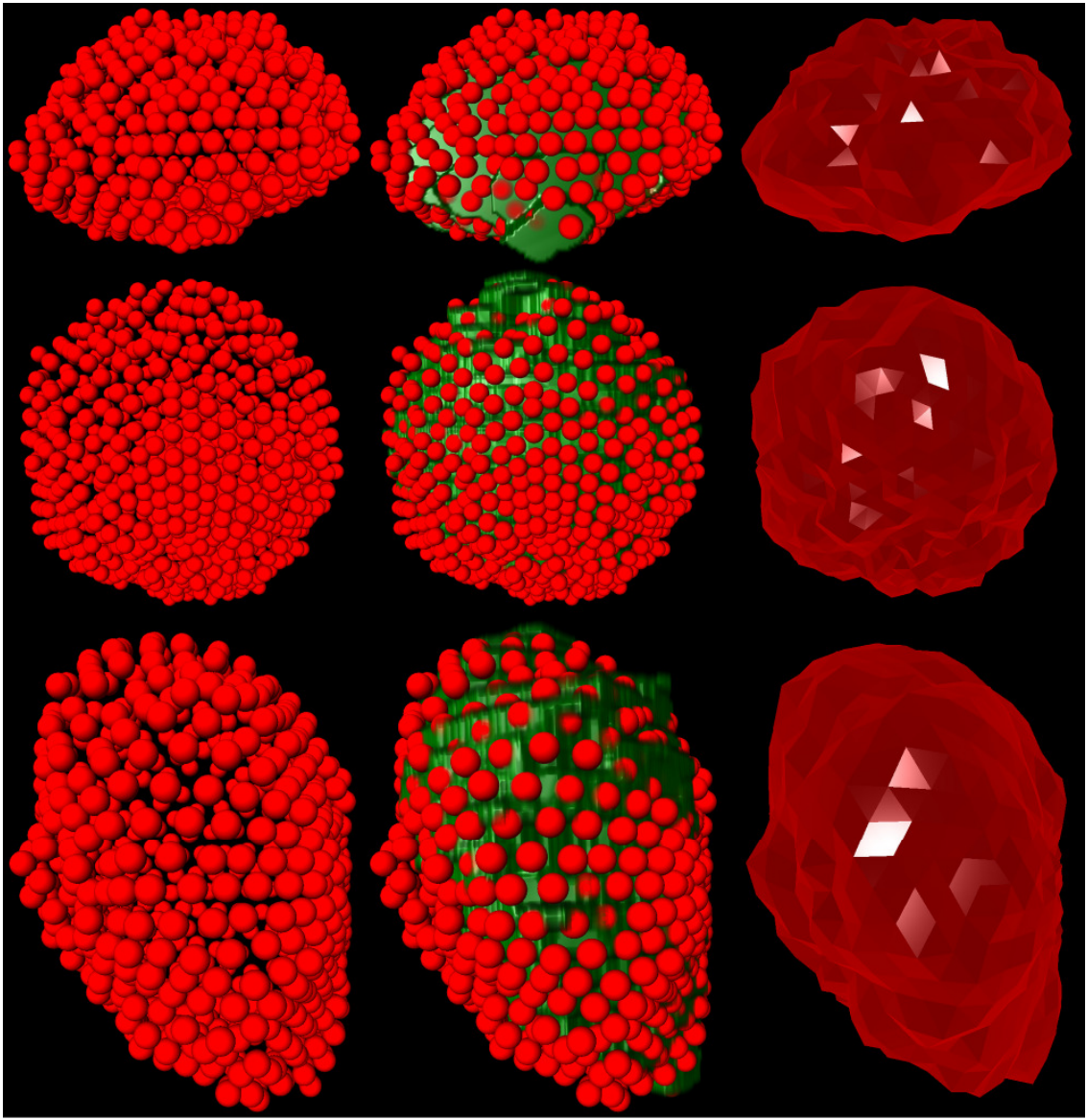

Figure 6

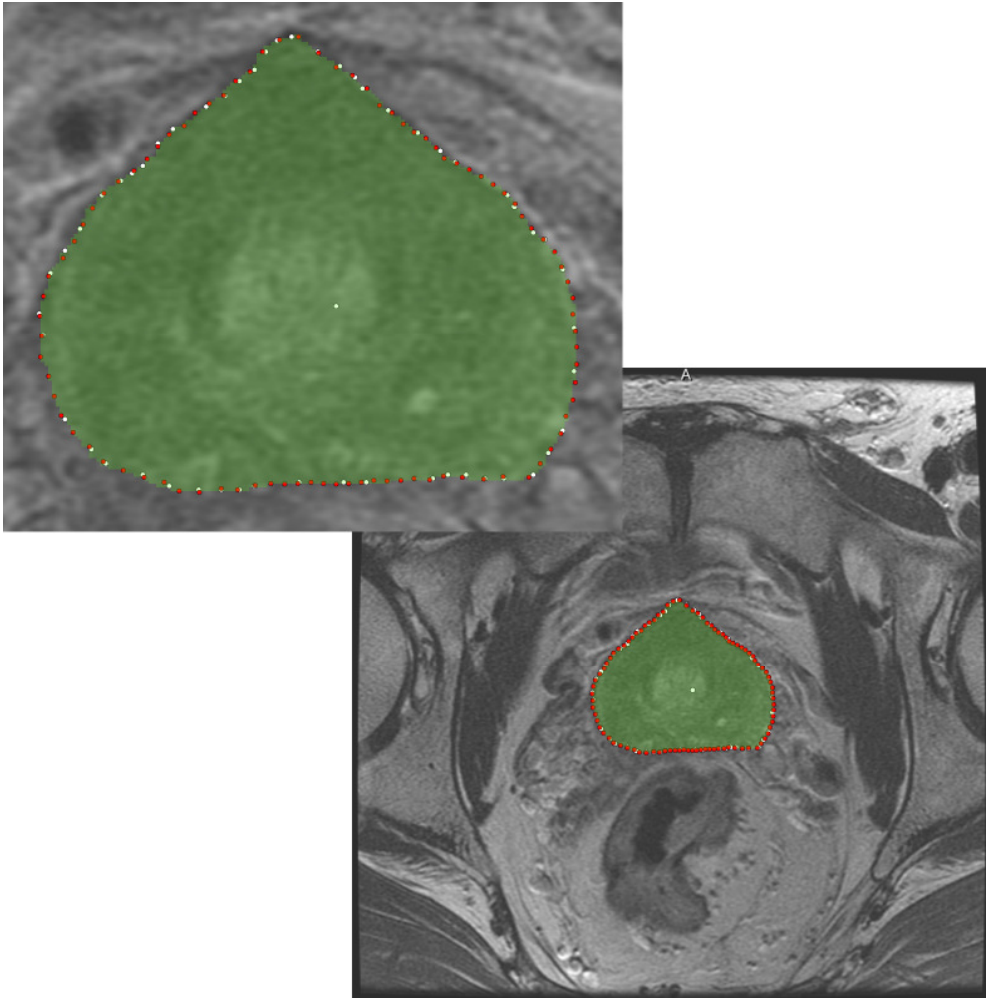